\pdfoutput=1
\documentclass[11pt]{article}

\usepackage[final]{acl}

\usepackage{times,latexsym}

\usepackage[T1]{fontenc}
\usepackage[utf8]{inputenc}

\usepackage{microtype,inconsolata}
\usepackage{multirow}
\usepackage{graphicx,booktabs}
\usepackage{hyperref, amsfonts, amsmath, enumitem}
\usepackage{algorithm}
\usepackage{algpseudocode}

\title{Prefixing Attention Sinks can Mitigate Activation Outliers\\for Large Language Model Quantization}

\author{
	Seungwoo Son\textsuperscript{\rm \texttt{1,2}}\thanks{Work done during internship at Google.},
	Wonpyo Park\textsuperscript{\rm \texttt{2}},
    Woohyun Han\textsuperscript{\rm \texttt{2}},
	Kyuyeun Kim\textsuperscript{\rm \texttt{2}},
	Jaeho Lee\textsuperscript{\rm \texttt{1,2}\thanks{Contact: \texttt{jaeho.lee@postech.ac.kr}}}\\
	\textsuperscript{\rm \texttt{1}}POSTECH\qquad \textsuperscript{\rm \texttt{2}}Google \\
    \texttt{\{swson, jaeho.lee\}@postech.ac.kr} \\
	\texttt{\{wppark, woohyun, kyuyeunk\}@google.com} \\
}
\usepackage{xspace}
\makeatletter
\DeclareRobustCommand\onedot{\futurelet\@let@token\@onedot}
\def\@onedot{\ifx\@let@token.\else.\null\fi\xspace}
\def\eg{\emph{e.g}\onedot} 
\def\ie{\emph{i.e}\onedot}

\makeatother

\algrenewcomment[1]{\(\triangleright\) #1}
\algnewcommand{\LineComment}[1]{\State \(\triangleright\) #1}

\DeclareMathOperator*{\argmin}{arg\,min} 

\definecolor{BrickRed}{rgb}{0.6,0,0}
\definecolor{BrickBlue}{rgb}{0,0,0.6}
\definecolor{BrickGreen}{rgb}{0,0.6,0}

\usepackage{colortbl}
\usepackage[capitalise]{cleveref}
\usepackage{csquotes}
\newcommand{\gr}{\rowcolor[gray]{.95}}

\begin{document}
\maketitle

\begin{abstract}
Despite recent advances in LLM quantization, activation quantization remains to be challenging due to the activation outliers. Conventional remedies, \eg, mixing precisions for different channels, introduce extra overhead and reduce the speedup. In this work, we develop a simple yet effective strategy to facilitate per-tensor activation quantization by preventing the generation of problematic tokens. Precisely, we propose a method to find a set of key-value cache, coined \textit{CushionCache}, which mitigates outliers in subsequent tokens when inserted as a prefix. CushionCache works in two steps: First, we greedily search for a prompt token sequence that minimizes the maximum activation values in subsequent tokens. Then, we further tune the token cache to regularize the activations of subsequent tokens to be more quantization-friendly. The proposed method successfully addresses activation outliers of LLMs, providing a substantial performance boost for per-tensor activation quantization methods. We thoroughly evaluate our method over a wide range of models and benchmarks and find that it significantly surpasses the established baseline of per-tensor W8A8 quantization and can be seamlessly integrated with the recent activation quantization method.

\end{abstract}

\section{Introduction}\label{sec:introduction} 
Tremendous capabilities of large language models (LLMs) come with a tremendous computational cost. Modern language models often have over hundreds of billions of parameters, requiring significant memory and computation for prediction and training. For instance, OPT-175B \citep{zhang2022opt}, one of the most popular open-sourced language models, requires at least 350GB of memory and the order of $10^{18}$ floating point operations to generate a new token\footnote{assuming the context length $2048$} \citep{hoffman22}.

Quantization is an effective strategy to reduce the computational cost of LLMs. Recent works demonstrate that the precision of LLM weight parameters can be greatly reduced by post-training quantization (PTQ), with minimal degradations in its generation quality. For example, \citet{huang2024good} shows that one can quantize the weights of the LLaMA3-70B to 4 bits, with less than 0.5\%p drop in its zero-shot prediction accuracy. Roughly, the reduced precision translates into 4$\times$ increase in the generation throughput, and similar reduction in memory requirements \citep{lin24}.

LLM activations, however, remain challenging to be quantized. The key obstacle is the activation outlier, \textit{i.e.,} a small number of activations that are substantially larger than others \citep{bondarenko21,dettmers22,sun2024massive}. Such outliers elongate the quantization range and flattens out most non-outlier activations, leading to large performance losses even at W8A8 quantization.

\begin{figure*}[t]
\centering
\includegraphics[width=0.95\textwidth]{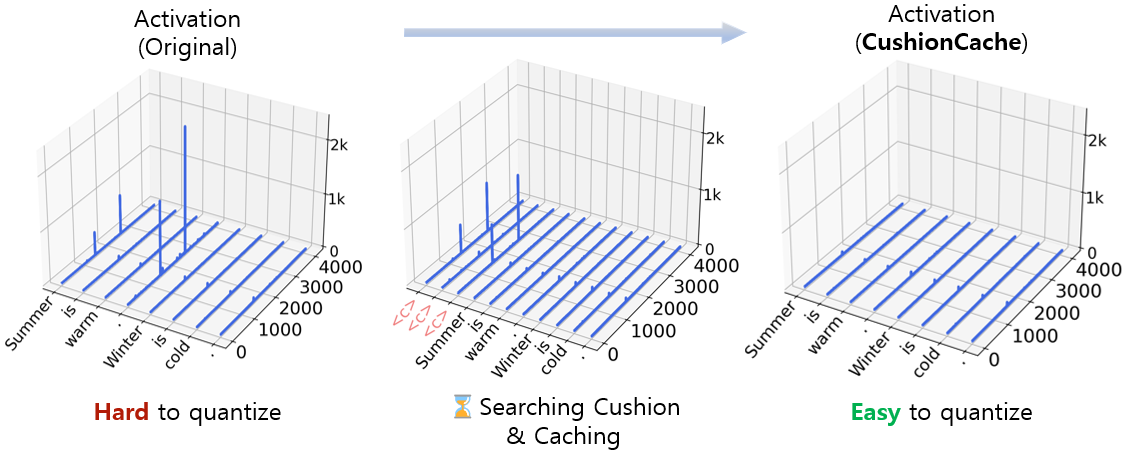}  \caption{\textbf{Activation magnitudes in LLaMA2-7B, before and after CushionCache.} CushionCache mitigates the activation outliers in LLMs by inserting and tuning the several prefix tokens to the model, which acts as an attention sink. Adding such sink tokens alleviates outliers in the subsequent tokens and enables a better activation quantization of the model with coarse quantization granularities.}
\vspace{-0.5em}
\end{figure*}

To address this issue, recent works propose to mitigate outliers based on various relaxations of the stringent \textit{static, per-tensor} quantization. One line of work applies quantization separately to each channel depending on the outlier proneness \citep{bondarenko21,dettmers22}. These methods, however, are difficult to be implemented on conventional hardwares. Another line of work reparameterizes the activations and weights in a way that the impact of outliers are amortized \citep{xiao23,ashkboos2024quarot}. These algorithms focuses on attaining high generative quality by adopting per-token or dynamic quantization, leaving the most hardware-friendly option---static per-tensor quantization---less explored.

To fill this gap, we take a novel approach for mitigating activation outliers in LLMs. In particular, we focus on answering the following key question:
\begin{displayquote}
\vspace{-0.3em}
\textit{Can we find a good \textbf{prefix} that mitigates the activation outliers in the subsequent tokens on a pretrained LLM?}
\vspace{-0.3em}
\end{displayquote}
Our answer is positive; we develop a very simple yet effective method, coined \textbf{\textit{CushionCache}}, to discover a prefix\footnote{more precisely, the key-value cache; we only care about the keys and values, rather than the token itself,} which reduces the outlier in the following tokens processed by the given LLM. By inserting this prefix, one then can quantize the activations of the LLM with much smaller quantization error, leading to an improved generation quality.

To design our method, we draw inspirations from a recent observation that the outliers may originate from \textit{attention sinks} \citep{bondarenko23}---the ``no-operation'' tokens that receive much attention from other tokens \citep{xiao2023efficient}. By adding sink-like tokens as a prefix, one may be able to separate out outlier activations as well, rendering the subsequent tokens outlier-free. In a nutshell, our method works in two steps. 
\begin{enumerate}[leftmargin=*,topsep=0.5pt,parsep=0.5pt,itemsep=0.5pt]
\item \underline{\textit{Greedy initialization.}} We search for a sequence of sink-like prompt tokens in a greedy manner, so that the activations of the subsequent tokens are less prone to outliers (\cref{ssec:greedy}).
\item \underline{\textit{Quantization-aware prefix tuning.}} We train the greedily initialized prefix further to minimize the combined loss of the prediction loss and quantization error (\cref{ssec:pretune}).
\end{enumerate}

Our experiments demonstrate that the proposed CushionCache is highly effective in making LLMs more quantizable. The technique is versatile, consistently improving the quantized performance of LLMs under various scenarios, from per-token to per-tensor static quantization. The method can also be seamlessly combined with existing quantization algorithms to further boost their performances. To summarize, we contribute the following.
\begin{itemize}
\item We introduce CushionCache, a new prefix discovery method for mitigating LLM outliers to improve the quantization performance. 
\item Through extensive experiments, we show that CushionCache can consistently improve the performance of quantized LLMs under a wide range of setup. In particular, we improve the prior state-of-the-art W8A8 per-tensor static range quantization of LLaMA3-8B over 30\%p in zero-shot accuracy on downstream tasks.
\item Through our analysis, we demonstrate that CushionCache effectively replaces the role of attention sink tokens.
\end{itemize}

\section{Related Work}\label{sec:related}

\paragraph{Outliers in LLMs.} The fact that there exists usually large entries in LLM activations, or outliers, has been reported by multiple works. \citet{kovaleva-etal-2021-bert} and \citet{timkey21} report the existence of outliers in large transformer-based language models (\eg, BERT), and find that they appear mostly in a small number of channels and layers. \citet{bondarenko21} make a similar observation in the context of quantization, and finds that quantizing the activations lead to a large degradation in generation quality; the work also reports that semantically meaningless tokens can have higher tendencies to have outliers. \citet{dettmers22} confirm the same finding while quantizing GPT-scale models, and studies how the model scale affects the prevalence of outliers over tokens and layers. More recently, \citet{sun2024massive} investigates a similar phenomenon in newer LLM variants and confirms that certain tokens are more likely to suffer from outliers.

\paragraph{Per-channel activation quantization.} A line of work proposes to mitigate outliers in LLM activation quantization by applying different scaling factors or precision to each channel. \citet{bondarenko21} splits activation channels into several groups and perform quantization on each group. LLM.int8() \citep{dettmers22} applies higher precision (\eg, FP16) to a small number of outlier-prone channels, while quantizing the other channels to lower bits (\eg, INT8). These works, however, are difficult to be implemented in conventional hardwares, as they requires scaling along the contracting dimension of matrix multiplication.

\paragraph{Per-token, with reparameterization.}
Another line of work proposes to quantize the activations per-token to reduce the impact of outliers with better hardware acceleration. Many of these works adopt \textit{reparameterization} of weights to mitigate the outliers further. ZeroQuant \citep{yao22} applies per-tensor quantization and knowledge distillation to achieve reasonable INT8 quantization performance. SmoothQuant \citep{xiao23}, Outlier Suppression+ \citep{wei23}, and OmniQuant \citep{shao2023omniquant} migrates the activation magnitudes to the weights to normalize the scales of the activations. More recently, QuaRot \citep{ashkboos2024quarot} rotates the activations so that the outlier magnitudes are distributed over multiple axes in the reparametrized space. While these methods are effective, per-token quantization are typically slower than per-tensor quantization at the same quantization precision as it requires larger scale which has a size of the number of tokens. 

\paragraph{Per-tensor quantization.}
Notably, \citet{xiao23} also provides two options for per-tensor activation quantization: one with dynamic quantization range, and another with static range.
While these options tend to be faster than per-token (with static range being the fastest), their generation quality is much lower than per-token, especially on recent models such as LLaMA3 \citep{touvron2023llama}.

\paragraph{Attention sinks and outliers.}
Recent works report an intriguing phenomenon in large transformers, termed \textit{attention sink}. \citet{xiao2023efficient} find that a small number of semantically meaningless tokens, usually at the beginning of the sequence, tend to receive unusually large attention. \citet{darcet2023vision} make a similar observation for vision transformers, and show that training ViTs with additional meaningless tokens can help make the attention structures more semantically meaningful. \citet{bondarenko23} hypothesize that the sink tokens may be the root cause of the activation outliers, and propose a new architecture that prevents the outliers from emerging when pretrained from scratch. Our work shares a similar intuition, but critically differs in that we mitigate outliers by fine-tuning the pretrained LLM. This means no modification to the network architecture is needed and does not need to train the model from scratch.

\section{Preliminaries}\label{sec:pre} 

\paragraph{Key-value cache.} Modern language models, typically based on decoder-only architecture, are built as a sequence of transformer blocks which process a sequence of tokens to predict the next token \citep{vaswani2017attention}. That is, at each decoding step, the transformer $f(\cdot)$ performs:
\begin{align}
    t_{n+1} = f(t_1,t_2,\ldots,t_n) \label{eq:ntp}
\end{align}
where $t_1,\cdots,t_n$ are the preceding tokens used as context. LLM iteratively applies \cref{eq:ntp} autoregressively to generate text as a sequence of tokens.

As the context length grows, the computational cost to process all previous tokens also grows larger, slowing down the generation significantly. A popular solution is to cache and reuse the keys and values of the preceding tokens computed during the previous iteration. This trick relies on the fact that preceding tokens affect the outcome of the current token only through their keys and values:
\begin{align*}
    (s_{n,1},\ldots,s_{n,n}) &= \mathrm{Attention}(q_n,k_{1:n})\\
    o_{n} &= \sum_{i=1}^n v_i \cdot s_{n,i}
\end{align*}
where $q_i, k_i, v_i$ denotes the query, key, and value vectors for each token and $s_{i,j}$ denotes the attention score for the $i$-th token query on $j$-th token key. By storing and reusing the cached values (called KV cache), one only needs to process new tokens as:
\begin{align}
    t_{n+l+1} = f(t_{n+1:n+l}~|~k_{1:n}, v_{1:n}),
\end{align}
where $t_{n+1:n+l}$ denotes $l$ tokens given at the current step and $k_{1:n}, v_{1:n}$ are the keys and values of the preceding context of length $n$, computed during the previous iteration. During the prefill phase, $l$ may be the length of the prompt, and during the decoding phase, we can simply use $l = 1$, processing only a single token at a time.

\paragraph{Quantization.} Quantization is an act of casting a high-precision tensor (typically FP) into a lower-precision tensor (typically INT), to save the memory to store and computation to process the tensor. In neural network quantization, a popular choice is the \textit{linear quantization}, which performs
\begin{align}
\mathbf{X}_{\mathrm{int}} = \mathrm{round}( (\mathbf{X}_{\mathrm{fp}}-z) / s ), \label{eq:int_quant_main}
\end{align}
where $\mathbf{X}_{\mathrm{int}}, \mathbf{X}_{\mathrm{fp}}$ denotes the quantized and original tensors, $z, s \in \mathbb{R}$ denote zero-point and scaling factor, and $\mathrm{round}(\cdot)$ denotes the rounding operation. The scaling factor is typically selected as
\begin{align}
s &= \frac{\max(\mathbf{X}_{\mathrm{fp}})-\min(\mathbf{X}_{\mathrm{fp}})}{2^{N-1} - 1}, \label{eq:int_quant_scaling}
\end{align}
where $N$ denotes the number of bits for the integer format. The zero-point is determined as either $z = \min(\mathbf{X}_{\mathrm{fp}})$ or $z=0$, for the asymmetric and symmetric quantization, respectively.

\paragraph{Activation quantization with static range.} By quantizing both activation and weight matrices, one can avoid performing computation-heavy FP matrix multiplications. That is, for the case of symmetric quantization (for simplicity), we approximate:
\begin{align}
    \mathbf{W}_{\mathrm{fp}}\mathbf{X}_{\mathrm{fp}} &\approx s_{\mathrm{W}}s_{\mathrm{X}} \cdot \mathbf{W}_{\mathrm{int}}\mathbf{X}_{\mathrm{int}},
\end{align}
where the right-hand side can be computed using an integer matrix multiplication, and a single multiplication of FP16/32 quantities (for scaling factors). The combined scaling factor need not be multiplied back to the matrix immediately, and can be used in the subsequent operations directly.

In many cases, the scaling factors $s_{\mathrm{W}}, s_{\mathrm{X}}$ can be pre-computed based on the validation set statistics. This method, called \textit{static-range quantization}, enables more acceleration than computing these values dynamically during the inference.

\paragraph{Outliers and complications.} In LLMs, the activation $\mathbf{X}_{\mathrm{fp}}$ tends to have a very large entry \citep{dettmers22,sun2024massive}. In such case, the magnitude of $\max(\mathbf{X}_{\mathrm{fp}})$ and $\min(\mathbf{X}_{\mathrm{fp}})$ will be very large, making the scaling factor $s_{\mathrm{X}}$ very large. This leads to a high sparsity in the tensor $\mathbf{X}_{\mathrm{int}}$, and a much degraded generation quality.

This problem can be alleviated in various ways: One can change the scaling factor dynamically over time (\ie, per-tensor dynamic quantization), or apply different scaling factors for each channel or token (\ie, per-channel/token quantization). As these methods require on-the-fly computations of scaling factors, the methods are typically slower.

\paragraph{Granularity and the communication cost.}
The drawback of finer quantization granularity becomes more significant in the distributed setup, as it affect the communication cost between nodes.

To see this, consider the case of multiplying matrices with tensor parallelism, \eg, Megatron-LM \citep{shoeybi2019megatron}. Comparing with the per-tensor static quantization, per-tensor dynamic quantization requires an additional AllReduce operation over the nodes to aggregate the (high-precision) scaling factor. The overhead is even more significant for per-token dynamic quantization, as the number of scaling factors is multiplied by the number of tokens, increasing the cost of AllReduce.

\section{Method}\label{sec:method}
We now describe CushionCache, an algorithm to find a prefix which can mitigate activation outliers in the subsequent tokens, thereby alleviating the quality degradation from activation quantization.

CushionCache aims to find a set of prefix that minimizes the quantization error of the activations. More concretely, let $\mathbf{X}_{i}$ denote activation of a transformer block for the input token $t_i$. Our goal is to minimize the squared difference between the original and the quantized activations, \ie,
\begin{align}
    &L_\text{q}(t_1, ..., t_n) = \sum^{n}_{i=1} \|\mathbf{X}_{i} - q(\mathbf{X}_{i})\|^2_2,
\end{align}
where $q(\cdot)$ denotes the quantization function, specified as $q(\mathbf{X}) =  s \cdot \mathrm{round}((\mathbf{X}-z)/s) + z$. In practice, we consider the summation of the error $L_q$ of all transformer blocks, but we omit this for the notational simplicity. Similarly, we define $L_{q}(t_{1:n}|p_{1:m})$ as the sum of squared error for $t_{1:n}$ given the prefix $p_{1:n}$, where the scaling factor $s$ and zero-point $z$ are determined for $t_{1:n}$ only.

We hypothesize that there exist prefix tokens $p$ that can reduce the expected activation quantization error of the tokens. That is, we find 
\begin{align}
\hat{p}_{1:m} = \argmin_{p_{1:m}}~ \mathbb{E}\left[ L_q(t_{1:n}~|~p_{1:m})\right], \label{eq:minimization}
\end{align}
where the expectation is taken over the probability distribution of tokens $t_{1:n}$. Once we find such prefix $\hat{p}_{1:m}$, their keys and values are cached and reused at the inference time to avoid redundant computation:
\begin{align}
t_{n+1} = f(t_{1:n}~|~\hat{k}_{1:m},~\hat{v}_{1:m}),
\end{align}
where $\hat{k}$ and $\hat{v}$ corresponds to the key-value caches of the prefix $\hat{p}_{1:n}$, which we call \textit{CushionCache}. 

We solve the minimization (eq.~\ref{eq:minimization}) with a strategy based on prefix tuning. This is done in two steps: Initializing prefixes based on greedily searched prompts (\cref{ssec:greedy}), and Quantization-aware Prefix tuning (\cref{ssec:pretune}).

\subsection{Greedy Prefix Search}\label{ssec:greedy}
We carefully initialize the prefix as the prefix tuning is known to be very sensitive to initial values. We follow \citet{li21} to search for the prefix that are activations of hard prompt tokens, \ie, input tokens that correspond to real text. As the search complexity grows exponentially with respect to the embedding size, we propose to use a greedy search algorithm with tailored heuristics.

In a nutshell, our method is a greedy search with \textit{\textbf{early stopping}}. We add new tokens to the prompt one-by-one, selected to minimize the quantization error. If the new token does not decrease the error much, we stop adding to prevent overfitting and computational overhead from long prompts.

Concretely, at each step, we first draw a single sample text $t_{1:n}$ from the dataset; we use the C4 dataset \citep{raffel2020exploring}, which is commonly used for calibration or validation purposes, to draw a sentence of length $n=512$. Then, based on the current state of prompts $p_{1:k}$, we search for the next prompt token $p_{k+1}$ by solving
\begin{align}
    p_{k+1} = \argmin_{p \in \mathcal{E}} L_q(t_{1:n}|p_{1:k},p),\label{eq:greedy_min}
\end{align}
where $\mathcal{E}$ denotes the embedding table; we can solve this problem rapidly by \textit{batched inference}. If the discovered new token reduces the quantization error by some fraction $\tau > 0$, \ie, satisfies
\begin{align}
    L_q(t_{1:n}|p_{1:k+1}) < \tau \cdot L_q(t_{1:n}|p_{1:k}),
\end{align}
then we append this token to the prompt and proceed to the next iteration. Otherwise, or if the max length is met, we stop searching for a new token. We use $\tau = 0.5$ for all experiments, which consistently shows a good performance.

Note that this algorithmic design provides some flexibility. More specifically, one can initialize the prompt with nonempty sequence before the search, as a heuristic that can help speed up the prompt search procedure. We find that filling in nonsemantic words, \eg, \texttt{<bos>} or \texttt{\textbackslash n}, is particularly useful; this observation is well-aligned with the findings of \citet{bondarenko21,sun2024massive}.

\begin{algorithm}[t]
\caption{Greedy prefix search}\label{alg:prefix_greedy_search}
\begin{algorithmic}[1]
\Require validation dataset $D$, embedding table $\mathcal{E}$, max length $m$, threshold $\tau$
\State $p = [\:]$
\hfill \Comment{{\color{gray}initialize the prompt}} 
\While{$\mathtt{len}(p) < m$}
\State $t \sim \mathrm{Unif}(D)$ \hfill \Comment{{\color{gray}draw a text}}
\State $p^{*}=\argmin_{p'\in\mathcal{E}} L_q(t|p,p')$. 
\If{$L_q(t|p,p^*) > \tau \cdot L_q(t|p)$}
\State \textbf{break}
\EndIf
\State $p.\mathtt{append}(p^*)$ \hfill \Comment{{\color{gray}add new token}}
\EndWhile
\State \Return $p$
\end{algorithmic}
\end{algorithm}

\begin{table*}[ht]
\small
\centering
\resizebox{0.9\textwidth}{!}{%
    
\begin{tabular}{l|lllll}
\toprule
\textbf{WikiText-2 ($\downarrow$)}          & \textbf{LLaMA2-7B} & \textbf{LLaMA3-8B} & \textbf{Mistral-7B-v0.1} & \textbf{OPT-6.7B}    & \textbf{BLOOM-7B}    \\
\midrule
FP16 & 5.47 & 6.13 & 5.25 & 10.86 & 11.37 \\
\midrule
Per-tensor Static & 9250.33 & 9759.46 & 85.51 & 11.45 & 11.93 \\
\gr + CushionCache (ours) & 5.98 {\color{BrickGreen}(-99.9\%)} & 7.41 {\color{BrickGreen}(-99.9\%)} & 5.84 {\color{BrickGreen}(-93.2\%)} & 11.00 {\color{BrickGreen}(-3.9\%)} & 11.50 {\color{BrickGreen}(-3.6\%)} \\
SmoothQuant-O3 & 15439.73 & 14022.91 & 618.27 & 10.85 & 11.55 \\
\gr + CushionCache (ours) & 5.87 {\color{BrickGreen}(-99.9\%)} & 7.37 {\color{BrickGreen}(-99.9\%)} & 5.60 {\color{BrickGreen}(-99.1\%)} & 10.68 {\color{BrickGreen}(-1.6\%)} & 11.38 {\color{BrickGreen}(-1.5\%)} \\
\midrule 

Per-tensor Dynamic & 8.01 & 23.86 & 67.86 & 11.73 & 11.81 \\
\gr + CushionCache (ours) & 5.69 {\color{BrickGreen}(-29.0\%)} & 7.30 {\color{BrickGreen}(-69.4\%)} & 5.59 {\color{BrickGreen}(-91.8\%)} & 10.99 {\color{BrickGreen}(-6.3\%)} & 11.43 {\color{BrickGreen}(-3.2\%)} \\
SmoothQuant-O2 & 8.13 & 25.12 & 66.16 & 10.87 & 11.59 \\
\gr + CushionCache (ours) & 5.66 {\color{BrickGreen}(-30.4\%)} & 7.29 {\color{BrickGreen}(-71.0\%)} & 5.56 {\color{BrickGreen}(-91.6\%)} & 10.68 {\color{BrickGreen}(-1.7\%)} & 11.39 {\color{BrickGreen}(-1.7\%)} \\
\midrule

Per-token Dynamic & 5.47 & 6.22 & 5.30 & 11.20 & 11.47 \\
\gr + CushionCache (ours) & 5.37 {\color{BrickGreen}(-1.8\%)} & 6.15 {\color{BrickGreen}(-1.1\%)} & 5.21 {\color{BrickGreen}(-1.7\%)} & 10.77 {\color{BrickGreen}(-3.8\%)} & 11.37 {\color{BrickGreen}(-0.9\%)} \\
SmoothQuant-O1 & 5.49 & 6.19 & 5.27 & 10.86 & 11.38 \\
\gr + CushionCache (ours) & 5.36 {\color{BrickGreen}(-2.4\%)} & 6.15 {\color{BrickGreen}(-0.6\%)} & 5.20 {\color{BrickGreen}(-29.0\%)} & 10.67 {\color{BrickGreen}(-1.7\%)} & 11.35 {\color{BrickGreen}(-0.3\%)} \\ 
\bottomrule                  
\end{tabular}
}
\vspace{-0.5em}
\caption{\textbf{Perplexity of W8A8-quantized LLMs on raw-WikiText2.} {\color{BrickGreen}Green} denotes the relative decrease.}\label{tab:wikitext}
\vspace{-0.5em}
\end{table*}

\begin{table*}[ht]
\small
\centering

\resizebox{0.9\textwidth}{!}{%

\begin{tabular}{l|lllll}
\toprule
\textbf{7 Zero-shot Tasks ($\uparrow$)}           & \textbf{LLaMA2-7B} & \textbf{LLaMA3-8B} & \textbf{Mistral-7B-v0.1} & \textbf{OPT-6.7B}    & \textbf{BLOOM-7B}    \\ \midrule
FP16 & 65.63 & 68.83 & 69.14 & 60.50 & 56.20                \\ \midrule
Per-tensor Static & 36.37 & 35.86 & 48.83 & 57.94 & 55.87 \\
\gr + CushionCache (ours) & 64.47 {\color{BrickGreen}(+28.10)} & 67.85 {\color{BrickGreen}(+31.99)}& 67.75 {\color{BrickGreen}(+18.91)}& 59.85 {\color{BrickGreen}(+1.91)}& 55.91 {\color{BrickGreen}(+0.04)}\\
SmoothQuant-O3 & 36.32 & 36.22 & 37.45 & 60.61 & 55.96 \\
\gr + CushionCache (ours) & 64.67 {\color{BrickGreen}(+28.35)} & 66.99 {\color{BrickGreen}(+30.77)} & 68.39 {\color{BrickGreen}(+30.94)} & 60.87 {\color{BrickGreen}(+0.26)}& 56.66 {\color{BrickGreen}(+0.75)}\\
\midrule

Per-tensor Dynamic & 61.94 & 58.94 & 52.02 & 59.23 & 56.46 \\
\gr + CushionCache (ours) & 65.34 {\color{BrickGreen}(+3.40)} & 68.66 {\color{BrickGreen}(+9.72)}& 69.02 {\color{BrickGreen}(+17.00)}& 60.28 {\color{BrickGreen}(+1.05)}& 58.47 {\color{BrickGreen}(+2.01)}\\
SmoothQuant-O2 & 61.24 & 58.67 & 51.08 & 60.57 & 56.14 \\
\gr + CushionCache (ours) & 65.65 {\color{BrickGreen}(+4.41)} & 68.74 {\color{BrickGreen}(+10.07)} & 69.15 {\color{BrickGreen}(+18.07)} & 60.60 {\color{BrickGreen}(+0.03)}& 58.99 {\color{BrickGreen}(+2.85)}\\
\midrule
Per-token Dynamic & 65.43 & 68.92 & 68.90 & 59.48 & 56.55 \\
\gr + CushionCache (ours) & 65.78 {\color{BrickGreen}(+0.35)} & 68.58 {\color{BrickRed}(-0.34)} & 69.83 {\color{BrickGreen}(+0.93)} & 60.65 {\color{BrickGreen}(+1.17)}& 56.72 {\color{BrickGreen}(+0.17)}\\
SmoothQuant-O1 & 65.64 & 68.64 & 69.09 & 60.55 & 56.35 \\
\gr + CushionCache (ours) & 65.97  {\color{BrickGreen}(+0.33)} & 68.78 {\color{BrickGreen}(+0.14)}& 69.99 {\color{BrickGreen}(+0.90)}& 61.01 {\color{BrickGreen}(+0.46)}& 56.80 {\color{BrickGreen}(+0.45)}\\
\bottomrule                  
\end{tabular}
}
\vspace{-0.5em}
\caption{\textbf{Average zero-shot accuracies of W8A8-quantized LLMs.} We average over LAMBADA, HellaSwag, PIQA, WinoGrande, OpenBookQA, RTE, and COPA. {\color{BrickGreen}Green} is the accuracy gain and {\color{BrickRed}red} is the drop.}\label{tab:zero_shot}
\vspace{-0.5em}
\end{table*}

\subsection{Quantization-aware Prefix Tuning} \label{ssec:pretune}
Using the intermediate activations of the greedily-searched prompt as an initial prefix, we fine-tune the CushionCache via prefix tuning \citep{li21}. Precisely, we freeze the model parameters and train the prefix with the loss
\begin{align}
L = L_{\mathrm{pred}} + \lambda \cdot L_q
\end{align}
where $L_{\mathrm{pred}}$ is the cross entropy loss for the next-token prediction and $\lambda$ is a hyperparameter that balances two losses. Here, we apply stop-grad to scaling factors and zero-points of the quantization function, as is typical in quantization-aware training literature \citep{jacob2018quantization}.

By optimizing this loss function, we ensure that the CushionCache not only improves the prediction accuracy but also minimizes the quantization error. This tuning does not require excessive amount of memory, as we only train the prefix.

\section{Experiments}\label{sec:experiment} %
\subsection{Experimental Setup}
\paragraph{Models.}
We evaluate our method on five LLM models: LLaMA2 and 3 \citep{touvron2023llama}, Mistral \citep{jiang2023mistral}, OPT \citep{zhang2022opt} and BLOOM \citep{workshop2022bloom}. 

\paragraph{Datasets.}
We measure the perplexity on the held-out set of WikiText-2 validation dataset \citep{merity2016pointer}. For zero-shot evaluation, we use seven tasks from the LM evaluation harness benchmark by EleutherAI \citep{eval-harness}. Precisely, we use LAMBADA, HellaSwag, PIQA, WinoGrande, OpenBookQA, RTE, and COPA datasets.

\paragraph{Base algorithms.} We apply CushionCache on two base activation quantization algorithms: Na\"{i}ve activation quantization and SmoothQuant \citep{xiao2023efficient}. We consider three different scenarios: Per-tensor static, per-tensor dynamic, and per-token dynamic quantization. Note that for each case, the SmoothQuant has a corresponding version, called O3, O2, and O1, respectively.

\paragraph{Configuration: Quantization.} We mostly follow the setup of \citet{li2024evaluating} and the TensorFlow default. We use symmetric group-wise quantization for model weights, and asymmetric quantization for the activations. For SmoothQuant, we use the migration strength $\alpha=0.8$, which worked consistently well throughout our experiments.  For static range quantization, we calibrate using the training split of WikiText-2 \citep{merity2016pointer}.

\paragraph{Configuration: Prefix tuning.} We follow the setup of \citet{li21} and tune for 2 epochs. We set the hyperparameter $\lambda = 0.01$.

\subsection{Main Results: W8A8 Quantization}
In \cref{tab:wikitext,tab:zero_shot}, we provide the performance achieved by the quantized language models, quantized with and without the proposed CushionCache. We report the WikiText perplexity and zero-shot accuracy in the tables, respectively.

For per-tensor static range quantization, CushionCache successfully improves the performance of the model; the boost is quite substantial in LLaMA and Mistral, often providing over 30\%p gains in terms of zero-shot accuracies. Intriguingly, the gain is much more pronounced in LLaMA-style models, which adopt the pre-LayerNorm and gated linear units. For per-tensor dynamic range quantization, similarly, we make consistent improvements over both vanilla quantization and SmoothQuant.

For per-token dynamic quantization, the gain is somewhat marginal, as the base quantization algorithms already tend to achieve a close performance to the FP16 model; we revisit per-token case for lower precision in \cref{ssec:lowerbit}.

\subsection{Ablation Study}
In \cref{tab:per_step}, we sequentially add our key algorithmic components to validate their efficacy. In particular, the components are (1) greedy-searched initial value, (2) prefix tuning, and (3) the quantization-error-based regularizer.

We observe that each component makes nontrivial contributions for achieving near-FP16 zero-shot accuracy. Interestingly, we find that the greedy-searched initialization is especially effective, contributing $\sim$91\% of the accuracy gain. This suggests that our search mechanism can be used as a compute-light standalone method in the cases where it is difficult to conduct prefix-tuning, due to a limited on-device memory.

\begin{table}[t]
    \small
    \centering
    \resizebox{0.45\textwidth}{!}{%
    
\begin{tabular}{ll} \toprule
\textbf{LLaMA3-8B}                                 & \textbf{Zero-shot acc. (\%)} \\ \midrule
FP16                                   & 68.83           \\ \midrule
Per-tensor Dynamic                             & 58.94           \\
+ Greedy-searched init.  & 67.78 {\color{BrickGreen}(+8.84)} \\
+ Prefix tuning                        & 68.13 {\color{BrickGreen}(+0.35)} \\
+ Quantization-aware loss & 68.66 {\color{BrickGreen}(+0.53)} \\ 
\bottomrule
\end{tabular}}
\vspace{-0.5em}
\caption{\textbf{Ablation study.} We compare the contribution of each algorithmic component by sequentially adding them. We apply W8A8 per-tensor dynamic quantization on the LLaMA3-8B model.} \label{tab:per_step}
\vspace{-0.5em}
\end{table}

\begin{table}[t]
    \small
    \centering
    \resizebox{0.48\textwidth}{!}{%
\begin{tabular}{llll}
\toprule
\textbf{Per-Token Dyn.} & \textbf{Perf.} & \textbf{LLaMA3-8B} & \textbf{Mistral-7B} \\ \midrule
& ppl \tiny{($\downarrow$)} & 6.13      & 5.25       \\
\multirow{-1.8}{*}{FP16}  & acc.\tiny{($\uparrow$)} &  68.83  & 69.14   \\ \midrule
SmoothQuant-O1 & ppl  & 6.93      & 5.49       \\
{\color{gray}(W6A6)} & acc.   &  66.72  & 67.51   \\
\rowcolor[gray]{.95} 
\cellcolor[gray]{.95}                                  & ppl   & 6.74 {\color{BrickGreen}(-2.7\%)} & 5.40 {\color{BrickGreen}(-1.6\%)} \\
\rowcolor[gray]{.95} 
\multirow{-1.8}{*}{\cellcolor[gray]{.95}+ CushionCache} & acc.   &  67.60 {\color{BrickGreen}(+0.88)} & 68.42 {\color{BrickGreen}(+0.91)} \\ \midrule
SmoothQuant-O1  & ppl   & 130.32    & 18.57      \\ 
{\color{gray}(W4A4)} & acc.   & 40.25   & 51.11       \\
\rowcolor[gray]{.95} 
\cellcolor[gray]{.95}                                  & ppl   & 29.09 {\color{BrickGreen}(-77.7\%)}  & 12.45 {\color{BrickGreen}(-33.0\%)}\\
\rowcolor[gray]{.95} 
\multirow{-1.8}{*}{\cellcolor[gray]{.95}+ CushionCache} & acc.   & 48.78 {\color{BrickGreen}(+8.53)} &    55.58 {\color{BrickGreen}(+4.47)} \\
\bottomrule
\end{tabular}}
\vspace{-0.5em}
    \caption{\textbf{W6A6/W4A4 quantization.} We additionally evaluate per-token quantization with lower bits, as W8A8 does not degrade much performance in general.} \label{tab:lower_bit}
\vspace{-0.5em}
\end{table}

\subsection{4/6-bit Per-token Quantization}\label{ssec:lowerbit}
To confirm the effectiveness of CushionCache under per-token dynamic quantization, we additionally evaluate with a lower precision (\cref{tab:lower_bit}). In particular, we use W6A6 and W4A4.

The results confirm that the proposed CushionCache is also effective in boosting the quantization performance of per-token activation quantization algorithms. In particular, CushionCache helps keeping the accuracy degradation quite low ($\sim$1\%p) for W6A6 quantization of both LLaMA3 and Mistral.

\begin{table}[t]
\small
\centering

\resizebox{0.45\textwidth}{!}{%
\begin{tabular}{llll}
\toprule
 \textbf{Model} & \textbf{Top-1} & \textbf{Top 10\%} & \textbf{Median} \\ \midrule
LLaMA2-7B                                            &     2461.40          &    0.59           & 0.23              \\
\gr + CushionCache (ours)                                            &    25.83           & 0.59               & 0.24               \\ \midrule
LLaMA3-8B                                            & 288.32              &    0.16           & 0.06               \\
\gr + CushionCache (ours)                                            & 4.94              & 0.16              & 0.06               \\ \midrule
Mistral-7B-v0.1                                      & 352.05              & 0.12               & 0.04               \\ 
\gr + CushionCache (ours)                                            & 3.51          & 0.12              & 0.04              \\
\bottomrule
\end{tabular}
}
\vspace{-0.5em}
\caption{\textbf{Top-1, top 10\%, and the median activation magnitudes of three LLMs.} Here, we inspect the input activations to the last transformer block.} \label{tab:max_outlier}
\vspace{-0.5em}
\end{table}

\subsection{Other experiments}
In the \cref{app:others}, we provide additional experiments results on the following topics:
\begin{itemize}[leftmargin=*,topsep=0pt,parsep=0pt,itemsep=0pt]
    \item Evaluations on MMLU dataset (\cref{sapp:mmlu})
    \item Latency measurements (\cref{sapp:latency})
    \item Compatibility with other quantization methods (\cref{sapp:otherquant})
\end{itemize} 

\section{Analysis}\label{sec:analysis}

We now conduct a brief sanity check. In particular, we ask the following questions.
\begin{itemize}[leftmargin=*,topsep=0pt,parsep=0pt,itemsep=0pt]
    \item Did the outliers disappear? (\cref{ssec:measuring})
    \item Did the CushionCache really replace the role of attention sink? (\cref{ssec:attention_sink})
    \item Will it be computationally viable to run CushionCache for large models? (\cref{ssec:searchtime})
\end{itemize}

\begin{figure*}[t]
\centering
\includegraphics[width=0.95\linewidth]{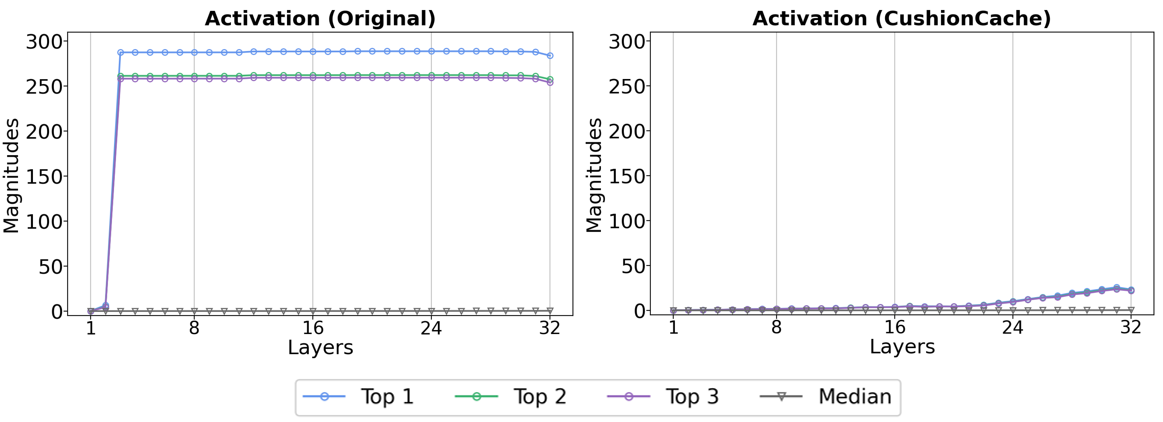}
\vspace{-0.5em}
\caption{\textbf{Top-1/2/3 and median activation magnitudes at each layer of LLaMA3-8B.} The left panel shows the activations without CushionCache, having significant outliers except for initial layers. The right panel shows the activation with CushionCache, having significantly reduced outliers in every layers.}  \label{fig:outlier_max}
\vspace{-0.5em}
\end{figure*}

\begin{figure*}[t]
\centering
\includegraphics[width=0.95\linewidth]{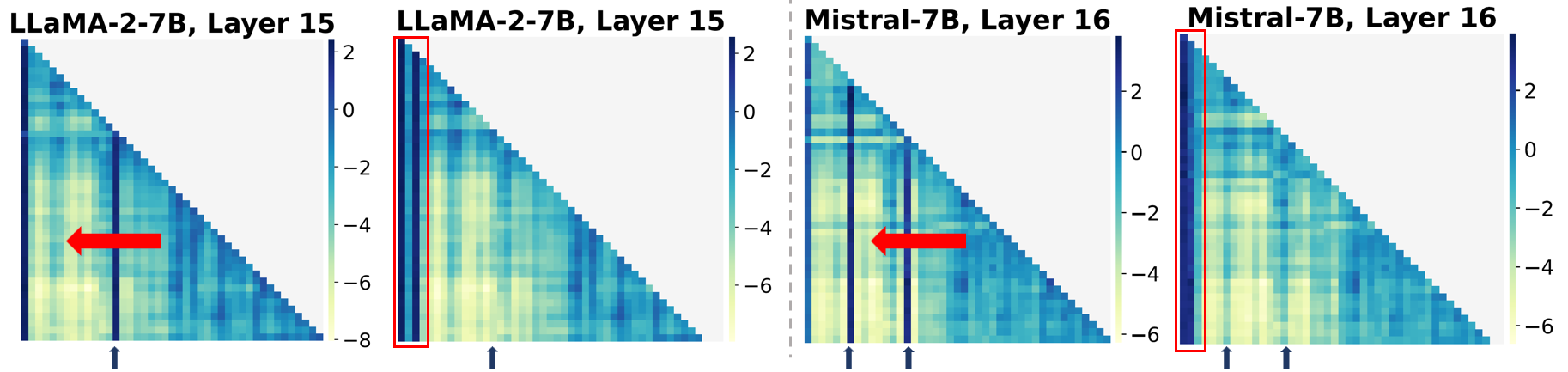}  
\vspace{-0.5em}
\caption{\textbf{Attention patterns before and after applying CushionCache in LLaMA3-8B and Mistral-7B.} The first and third panels show the attention patterns in models without CushionCache, where the attention sinks are quite prevalent in the generated token sequence. The second and fourth panels illustrate the attention patterns after inserting CushionCache. By adding the CushionCache, the attention is redirected toward the CushionCache tokens, preventing the attention sink from arising in the subsequent tokens.}  \label{fig:attention_sink}
\vspace{-0.5em}
\end{figure*}

\subsection{Change of Activation Magnitudes} \label{ssec:measuring}

In \cref{tab:max_outlier}, we report various order statistics of the activation magnitudes that appear in LLaMA2/3 and Mistral. In particular, we focus on the input activations to the last transformer block of these models, and measure the top-1, top 10\%, and median (\ie, top 50\%) activation magnitude. We average over ten samples, with a sequence length 4096.

The effect of CushionCache is quite dramatic. In particular, we find that the CushionCache can reduce the scale of the activation outlier to 1-2\% of the previous value. The ratio between the top-1 and the median decreases from roughly 10,000:1 to 100:1. We also note that the other order statistics, \ie, top 10\% and median, remains roughly the same before and after the CushionCache.

In \cref{fig:outlier_max}, we visualize the top-1/2/3 activations and median for each layer of LLaMA3-8B. The left panel plots the magnitude of the median and top-3 activations that occur during the standard operation of LLaMA3-8B. We observe that the median is almost zero, indicating that a significant fraction of all activations are close to zero, with only a few significantly large outliers. On the right panel, we plot the same values after applying the proposed CushionCache algorithm. We observe that the size of the top-3 activations have dramatically decreased, leading to a conclusion that CushionCache effectively removes the activation outliers. 

\subsection{Attention on CushionCache} \label{ssec:attention_sink}
In \cref{fig:attention_sink}, we visualize the attention patterns of LLaMA2 and Mistral, before and after applying the CushionCache. Attention sinks, as identified by \citet{xiao2023efficient,sun2024massive}, are tokens that disproportionately attract attention. By inserting CushionCache, we observe that the CushionCache tends to dominate most of the attention from other tokens, removing the sinks in other tokens.

\subsection{Time Needed to Search CushionCache}\label{ssec:searchtime}

In \cref{tab:search_cost}, we report the wall-clock time spent for performing the greedy search and prefix tuning of CushionCache. We observe that the greedy prefix search can be quite time-consuming, highly dependent on the side of the embedding table; LLaMA3-8B has a large embedding table. Another observation is that the quantization-aware prefix tuning step takes relatively small time for all models.

\begin{table}[t]
\centering
\small
\resizebox{0.45\textwidth}{!}{%
\begin{tabular}{lccc}
\toprule
\textbf{Model} & \textbf{Step 1} & \textbf{Step 2} & \textbf{Total Time} \\
\midrule
LLaMA2-7B & 2.68 hours & 3.34 hours & 6.02 hours \\
LLaMA3-8B & 12.09 hours & 3.70 hours & 15.79 hours \\
OPT-7B & 1.38 hours & 2.71 hours & 4.09 hours \\
\bottomrule
\end{tabular}
}
\vspace{-0.5em}
\caption{\textbf{Wall-clock time for the search.} We use a server with four NVIDIA A6000 GPUs.}
\label{tab:search_cost}
\vspace{-0.5em}
\end{table}

\section{Conclusion}\label{sec:conclusion} %
In this paper, we present CushionCache, a novel approach for mitigating activation outliers in LLMs to improve activation quantization performance. Through extensive experiments, we demonstrate that CushionCache consistently enhances the performance of per-tensor activation quantization. Our analysis shows that CushionCache effectively reduces the magnitude of activation outliers and redirects attention sinks, leading to more uniform and quantization-friendly activations. In contrast with other approaches to faciliate activation quantization, CushionCache is the first---up to our knowledge---to fundamentally alter the activation distribution itself without extensive training, making activations easier to quantize.

\section*{Limitations}
A limitation of our study is that our method is designed for LLMs with the decoder-only transformer structure. An extension to encoder-decoder LLMs \citep{raffel2020exploring} may require further modifications to the algorithm. Another limitation is the lack of a principled mechanism to determine the hyperparameter $\tau$, which decides when to stop adding new tokens. An extensive tuning may incur a non-negligible computational cost, especially when the target model is extremely large.

\section*{Ethics statement}
All experimental results we provide in this paper is based on publicly available datasets and open-source models, whose intended use include research purposes. We have used an AI assistant for the grammar check.

\section*{Acknowledgements}
This work was supported in part by the National Research Foundation of Korea (NRF) grant funded by the Korea government (MSIT) (No. RS-2023-00213710), and in part by the Korea government (MSIT) (No. RS-2024-00453301). JL also thanks Hong-Seok Kim and Radha Chandika for their generous support during his visit at Google.

\bibliography{main}
\clearpage
\newpage
\appendix
\section{Additional experiments}\label{app:others}
In this section, we provide additional experiments that have been missing in the main script.

\subsection{MMLU dataset}\label{sapp:mmlu}
We have additionally evaluated the quantized model on MMLU dataset, which encompasses a much larger set of tasks including STEM (\cref{tab:mmlu}). The results suggest that CushionCache remains to be effective on MMLU as well.

\begin{table}[h]
\small
\centering

\resizebox{0.45\textwidth}{!}{%
\begin{tabular}{llll}
\toprule
\textbf{Model} & \textbf{LLaMA2-7B} & \textbf{Mistral-7B} & \textbf{LLaMA3-8B} \\
\midrule
FP16 & 41.27 & 58.63 & 62.13 \\
\midrule
SmoothQuant-O3 & 23.76 & 23.62 & 25.32 \\
\gr + CushionCache (ours) & 38.06 {\color{BrickGreen}(+14.30)} & 56.59 {\color{BrickGreen}(+32.97)} & 58.99 {\color{BrickGreen}(+33.67)}\\
\midrule
SmoothQuant-O2 & 27.88 & 25.66 & 30.94 \\
\gr + CushionCache (ours) & 40.45 {\color{BrickGreen}(+12.59)} & 58.05 {\color{BrickGreen}(+32.39)} & 60.59 {\color{BrickGreen}(+29.65)} \\
\midrule
SmoothQuant-O1 & 40.76 & 58.70 & 61.88 \\
\gr + CushionCache (ours) & 41.65 {\color{BrickGreen}(+0.89)} & 59.20 {\color{BrickGreen}(+0.50)} & 61.55 {\color{BrickRed}(-0.33)}\\
\bottomrule
\end{tabular}
}
\vspace{-0.5em}
\caption{\textbf{Results on MMLU.} We compare the results on the MMLU dataset.} \label{tab:mmlu}
\vspace{-0.5em}
\end{table}

\subsection{Generation latency}\label{sapp:latency}
We have measured the average latency of generating each token. We have experimented with W8A8-quantized LLaMA-3B, using the SmoothQuant kernel on a single A6000 GPU; not that this may not be the best hardware-optimized kernel for our hardware, but can be meaningful in terms of providing a comparison. We have used the prompts of length 500, and averaged over 1000 generated tokens. We compare both the time to the first token (TTFT; prefill phase) and the time per output token (TPOT; generation phase). in \cref{tab:latency}.

We observe that CushionCache only adds negligible latency, while enabling a much better adoption of the per-tensor static-range quantization techniques which provides a much faster decoding. In particular, we observe that adding CushionCache adds only 
0.01--0.3ms in TTFT or TPOT, which is less than 0.5\% of the total latency. Furthermore, as the CushionCache makes the faster option (e.g., per-tensor static) a viable option, it can even be viewed as enabling an overall speedup up to a few milliseconds.

\begin{table}[h]
\small
\centering

\resizebox{0.45\textwidth}{!}{%
\begin{tabular}{lll}
\toprule
\textbf{LLaMA3-8B} & \textbf{TTFT (ms)} & \textbf{TPOT (ms)}\\
\midrule
Per-Tensor Static & 78.01 & 48.71 $\pm$ 1.58 \\
\gr + CushionCache (ours) & 78.22 & 48.86 $\pm$ 0.55 \\
\midrule
Per-Tensor Dynamic & 81.52 & 50.56 $\pm$ 0.71 \\
\gr + CushionCache (ours) & 81.56 & 50.93 $\pm$ 0.82 \\
\midrule
Per-Token Dynamic & 83.35 & 51.75 $\pm$ 1.22 \\
\gr + CushionCache (ours) & 83.64 & 51.76 $\pm$ 0.95 \\
\bottomrule
\end{tabular}
}
\vspace{-0.5em}
\caption{\textbf{Generation latency.} We measure the generation speed on LLaMA3-8B.} \label{tab:latency}
\vspace{-0.5em}
\end{table}

\newpage

\subsection{Other quantization methods}\label{sapp:otherquant}
We have conducted additional experiments to combine with quantization algorithms other than SmoothQuant. In particular, we have conducted experiments on the following recent methods:
\begin{itemize}[leftmargin=*,topsep=0pt,parsep=0pt]
\item AWQ-4bit \citep{lin24}: A recent weight-only quantization algorithm, for demonstrating that CushionCache effectively boosts the performance of weight quantization algorithms.
\item QuaRot-4bit \citep{ashkboos2024quarot}: A recent weight+activation+cache quantization algorithm, for demonstrating that CushionCache works well with SOTA quantization algorithms.
\item KIVI-2bit \citep{kivi}: A recent KV cache quantization algorithm, for demonstrating that the KV cache of the CushionCache-quantized model can be compressed well with KV cache quantization methods.
\end{itemize}
The results are given in \cref{tab:compatible}, where we observe that the CushionCache is indeed versatile, being able to be combined well with many different quantization methods. Note that for KIVI, we measure the GSM8K results, as the original paper does not report perplexity.

\begin{table}[h]
\small
\centering
\resizebox{0.45\textwidth}{!}{%
\begin{tabular}{ll} \toprule
\textbf{LLaMA3-8B} & \textbf{WikiText-2 Perplexity} \\ \midrule
FP16 & 6.13 \\ \midrule
AWQ & 6.18 \\
+ CushionCache  & 6.15 \\
+ Per-Cushion Static & 8.40 \\
+ Per-Cushion Static + CushionCache & 7.01 \\ 
\midrule
QuaRot & 8.21 \\
+ CushionCache  & 7.41 \\ \midrule
\midrule
\textbf{LLaMA3-8B} & \textbf{GSM8K (\%)} \\ \midrule
FP16 & 48.75 \\
+ KIVI & 42.61 \\ \midrule
Per-tensor Static & 0.06 \\
+ KIVI  & 0.03 \\
+ KIVI + CushionCache & 38.29 \\
\bottomrule
\end{tabular}}
\vspace{-0.5em}
\caption{\textbf{Other quantization methods.} We combine CushionCache with AWQ, QuaRot, and KIVI.} \label{tab:compatible}
\vspace{-0.5em}

\end{table}
\end{document}